\documentclass[journal]{IEEEtran}
\usepackage{bbm}
\usepackage{array}
\newcolumntype{P}[1]{>{\centering\arraybackslash}p{#1}}
\usepackage{algorithm}
\usepackage{algorithmicx}
\usepackage{algpseudocode}
\usepackage{tabu}
\usepackage{amssymb}
\usepackage{mathrsfs}
\usepackage{amsmath,cite}
\usepackage{chemarrow}
\usepackage{hyperref}
\usepackage{url}
\usepackage{color}
\usepackage{graphicx}          
\usepackage{cite} 
\usepackage{multirow}
\usepackage{amsmath}
\usepackage{epstopdf}
\newtheorem{prop}{Proposition}
\newtheorem{rem}{Remark}

\ifCLASSINFOpdf
\else
\fi
\begin{document}
%
\title{Bayesian Inference of Stochastic Dynamical Networks}
%
%
%

\author{Yasen~Wang, 
        Junyang~Jin, 
        and~Jorge Gon\c{c}alves
\thanks{This work is supported by the National Key R\&D Program of China (Grant number: 2018YFB1701202), and the Fundamental Research Funds for the
Central Universities (HUST: 2021JYCXJJ028). Corresponding author: Junyang Jin, email:  jj415@alumni.cam.ac.uk.}
\thanks{Yasen Wang is with the School of Mechanical Science and Engineering, and the State Key Lab of Digital Manufacturing Equipment and Technology, Huazhong University of Science and Technology,
Wuhan 430074, China.}
\thanks{Junyang Jin is with the HUST-Wuxi Research Institute, Wuxi 214174, China.}
\thanks{Jorge Gon\c{c}alves is with the Luxembourg Centre for Systems Biomedicine, University of Luxembourg, L-4367 Belvaux, Luxembourg, and with the Department of Plant Sciences, University of Cambridge, Cambridge, CB2 1TN, UK.}}

%
%

\markboth{}
{Shell \MakeLowercase{\textit{et al.}}: Bare Demo of IEEEtran.cls for IEEE Journals}
%



\maketitle

\begin{abstract}
Network inference has been extensively studied in several fields, such as systems biology and social sciences.   Learning network topology and internal dynamics  is essential to understand mechanisms of complex systems. In particular, sparse topologies and stable dynamics are fundamental features of many real-world continuous-time (CT) networks.  Given that usually only a partial set of nodes are able to observe, in this paper, we consider linear CT systems to depict networks since they can model unmeasured nodes via transfer functions. Additionally, measurements tend to be noisy and with low and varying sampling frequencies. For this reason, we consider CT models  since discrete-time approximations often require fine-grained measurements and uniform sampling steps.  The developed method applies dynamical structure functions (DSFs) derived from linear stochastic differential equations (SDEs) to describe networks of measured nodes.  A numerical sampling method, preconditioned Crank-Nicolson (pCN), is used to refine coarse-grained trajectories to improve inference accuracy. The convergence property of the developed method is robust to the dimension of data sources. Monte Carlo simulations indicate that the developed method outperforms state-of-the-art methods including group sparse Bayesian learning (GSBL), BINGO, kernel-based methods, dynGENIE3, GENIE3, and ARNI. The
simulations include random and ring networks, and a synthetic biological network. These are challenging networks, suggesting that the developed method can be applied under a wide range of contexts, such as gene regulatory networks, social networks, and communication systems.
\end{abstract}

\begin{IEEEkeywords}
System identification, stochastic dynamical systems, sparse network inference,  pCN sampling
\end{IEEEkeywords}

%
\IEEEpeerreviewmaketitle

\section{Introduction}
%
%
%
%
\IEEEPARstart{M}{ost} real-world networks evolve in continuous-time (CT), and have sparse topologies and stable internal dynamics. For example, gene regulatory networks are inherently stable and sparse: genes only regulate a small number of other genes.  Power grids are typically sparse to reduce energy loss during transmission and stable to maintain daily operation.  This paper addresses the inference of sparse and stable linear continuous-time networks from time series data.  One of the biggest challenges of network inference is to handle poor measurements. In systems biology, for example, data typically consists of low and non-uniform sampling frequencies. In addition, data are exposed to internal and external noise. Furthermore, usually only a partial set of nodes are measured (e.g. mRNA or proteins). Hence, methods for the  linear continuous-time network inference must explicitly take these limitations into account.

Linear systems have been widely used to approximate nonlinear behaviors of real-life networks. While linear systems are simpler than nonlinear systems, they are considerably easier to control, simulate and analyze. For example, linear systems can easily impose system stability and handle unmeasured nodes via transfer functions. Continuous-time (CT) models  capture dynamics of systems where rates can change at any moment in time, such as in biological systems. 
With fast sampling, discrete-time (DT) models can provide good approximations of CT.  However, modeling DT models from low sampling rates  as approximations of CT dynamics can easily lead to completely wrong results~\cite{yue}. Moreover, discretization of CT systems often demands uniform sampling. In practice, CT models are much harder to identify than DT.
 Aside from  discretizing models, one of the main techniques to identify CT models is to first estimate derivatives from time series~\cite{CTid}. However, these methods require fine-grained samples and are sensitive to measurement noise, which can lead to time-varying DT models and increase complexity dramatically~\cite{sysid}. In addition, these methods cannot handle systems with process noise. In general, preprocessing time series via, for example filtering or interpolation, may introduce bias to inferred networks~\cite{IB}. Hence, for low sampling frequencies, CT systems should ideally be identified directly without estimating derivatives or discretizing models. To this end, this paper incorporates trajectory refinement into network inference by numerical sampling. Refined trajectories are updated iteratively along with network topologies and system dynamics, which greatly improves inference accuracy.

Various methods have been developed to infer continuous-time networks including ARNI~\cite{ARNI}, dynGENIE3~\cite{dynGENIE3},  GENIE3~\cite{GENIE3}, kernel-based methods~\cite{nonp} and BINGO~\cite{BINGO}. ARNI is based on the greedy search and shows promising performance on different types of networks.
dynGENIE3 was compared with several state-of-the-art methods~\cite{dynGENIE3} and its minor variant, GENIE3, was the winner of a sub-challenge in DREAM4~\cite{dream4}. In recent years, kernel  methods  are combined with group sparse Bayesian learning (GSBL)  to identify linear systems~\cite{nonp,nonp1,nonp3}. They show several advantages in real-world applications. Most importantly, kernel-based methods do not require estimating model complexity and can impose system stability effectively~\cite{nonp1, nonp2}. Whilst these methods presented good performance on several examples, they are very sensitive to data quality. ARNI and dynGENIE3 rely on the quality of derivative estimation while GENIE3, and kernel-based methods~\cite{nonp} apply model discretization. Hence, all these methods require fine-grained measurements. While BINGO can address this issue remarkably and have shown superior performance on many different gene regulatory
networks, it is not designed for partially observed networks. 
We will return to these methods in the examples section and compare them with our proposed algorithm.
	
	In contrast to the above methods, this paper aims to model linear CT networks directly. It applies Gaussian processes~\cite{gauss} and Markov chain Monte Carlo (MCMC~\cite{pattern, mcmc}) methods to infer sparse continuous-time networks. Linear stochastic differential equations (SDEs~\cite{sto})  describe general networks, while dynamical structure functions (DSFs~\cite{qpmodel, yuan}) derived from linear SDEs encode unmeasured nodes via transfer functions. As a non-parametric method, Gaussian processes impose system stability and network sparsity. Then, MCMC methods simultaneously explore internal dynamics and network topology. In particular, preconditioned Crank-Nicolson sampler (pCN~\cite{pCN}) samples the true trajectories (no measurement noise) at finer time instances. By merit of pCN, the performance of the algorithm is robust to high-dimensional datasets.  Additionally, we introduce  sparse likelihood approximation techniques~\cite{sla,pseu} to reduce computational costs when dealing with large datasets.   In summary, the contribution of this paper is threefold;
\begin{itemize}
   \item The developed method is  capable of handling noisy time series that are sparsely and non-uniformly sampled. In addition,  network nodes are partially observed.
   
\item To address the issue that the Wiener processes are nowhere differentiable, we propose an equivalent realization of the  considered linear SDEs. As such, while only a partial set of nodes are measured, we are able to describe  network topologies of measured nodes by DSFs.
 
        \item Monte Carlo simulations show that the proposed method achieves  state-of-the-art performance compared with kernel-based methods, GSBL, BINGO, dynGENIE3, GENIE3 and ARNI on several challenging networks. Moreover, networks are inferred without parameter tuning.
\end{itemize}

The paper is organized as follows. Section \ref{Model} introduces dynamical structure function and formulates the full Bayesian model. Section~\ref{MCMC} applies MCMC methods to sample refined trajectories and model parameters. Section~\ref{Simulation} tests the proposed method under various experimental conditions. Finally, Section~\ref{Conclusion} concludes and discusses further development in this field.

	\emph{Notation}: This paper applies the matrix notation of Matlab. For matrix $A$, $A(i:j,p:q)$ presents elements from $i$th to $j$th rows and from $p$th to $q$th columns. Given matrices $A_1$ and $A_2$ of $\mathcal{R}^{m\times n}$, $[A_1,A_2]\in\mathcal{R}^{m\times2n}$ and $[A_1;A_2]\in\mathcal{R}^{2m\times n}$. Additionally, $I_n$ denotes a $n\times n$ identity matrix. For $A\in{\mathcal{R}}^{n\times n}$, $diag\{A\}$ denotes a vector which consists of the diagonal elements of matrix $A$. $[A]_{ij}$ presents the $ij$th entry of matrix $A$ while $V_i$ denotes the $i$th element of vector $V$. For a series of matrices, $\{A_i| i=1,...,n\}$, $blkdiag\{A_1,...,A_n\}$ denotes a block diagonal matrix.	 $\mathcal{N}(x|m,\Sigma)$ denotes the Gaussian distribution of random variable $x$ with mean $m$ and covariance matrix $\Sigma$. $U(x;a,b)$ denotes the Uniform distribution on $[a,b]$. $zeros(m,n)$ denotes a $m\times n$ matrix of zeros.

\section{MODEL FORMULATION}\label{Model}
\subsection{The dynamical structure function}
The sparse network of $n$ nodes is described by a linear stochastic differential equation (SDE~\cite{sto}) as follows:
\begin{equation}
\begin{aligned}
dx(t) &= [Ax(t)+Bu(t)]dt+KdW(t),\\
y(t) &= Cx(t),
\label{sde}
\end{aligned}
\end{equation}
	where $x\in \mathcal{R}^n$ are state variables of the system with  each representing a node in the network, $u\in \mathcal{R}^q$ are deterministic inputs, and $W\in \mathcal{R}^m$ denote independent Wiener processes whose increments are independent Gaussian random variables: for each row $r$, $W_r(t)-W_r(s)=\mathcal{N}(0,t-s)$.  Finally, $y\in \mathcal{R}^p$ present outputs of the system,  $A\in \mathcal{R}^{n\times n}$, $B\in \mathcal{R}^{n\times q}$, $K\in \mathcal{R}^{n\times m}$, and $C\in \mathcal{R}^{p\times n}$ are system matrices, typically unknown.
	
	In practice, not all the nodes are measurable. Usually, only a partial set of genes are measured in large-scale systems, such as genome-wide regulatory networks. Assuming only the first $p<n$ states are measured (i.e. $C=[I,\bold{0}]$), the goal is to infer partial networks of measured nodes. DSFs are able to encode unmeasured nodes via transfer functions~\cite{qpmodel}. Without process noise, DSFs are derived from state space models using the Laplace transformation. However, this approach cannot be applied to SDEs directly since Wiener processes are almost nowhere differentiable~\cite{sto}. To avoid this issue, we first propose an equivalent realization of model~\eqref{sde}.
\begin{prop} Given deterministic inputs and Wiener processes, the following state space model has an identical output as model~\eqref{sde}:
\begin{equation}
\begin{aligned}
\dot{x}(t) &= Ax(t)+Bu(t)+AKW(t),\\
y(t) &= Cx(t)+CKW(t).
\end{aligned}
\label{eqsde}
\end{equation}
\end{prop}	

	 \emph{Proof:} Since the Wiener process is continuous with probability $1$, there exists a unique solution to model~\eqref{eqsde} for almost all sample paths~\cite{ODE}. The solution to model~\eqref{eqsde} with initial state $x_{0}$ is:
\begin{equation}{\tiny
\begin{aligned}
y&=Ce^{At}x_{0}+\int_{0}^tCe^{A(t-\tau)}Bud\tau+CKW+\int_{0}^tCe^{A(t-\tau)}AKWd\tau\\
&=Ce^{At}x_{0}+\int_{0}^tCe^{A(t-\tau)}Bud\tau+CKW+\int_{0}^tCAe^{A(t-\tau)}KWd\tau.\\
\end{aligned}}
\end{equation}	
The solution to model~\eqref{sde} with initial state $x_{0}$ is:
\begin{equation}{\tiny
\begin{aligned}
y &= Ce^{At}x_{0}+\int_{0}^tCe^{A(t-\tau)}Bud\tau+\int_{0}^tCe^{A(t-\tau)}KdW\\
&=Ce^{At}x_{0}+\int_{0}^tCe^{A(t-\tau)}Bud\tau+CKW-\int_{0}^tCde^{A(t-\tau)}KW\\
&=Ce^{At}x_{0}+\int_{0}^tCe^{A(t-\tau)}Bud\tau+CKW+\int_{0}^tCAe^{A(t-\tau)}KWd\tau.\\
\end{aligned}}
\end{equation}	
where the second step uses the law of integral by parts~\cite{sto}.  This completes the proof. $\hfill\blacksquare$

To remove the hidden nodes from model~\eqref{eqsde}, state variables and system dynamics are divided as follows:
\begin{equation}
\begin{aligned}
 \left[\begin{array}{c}\dot{x}_1\\\dot{x}_2\end{array}\right]&=  \left[\begin{array}{cc}A_{11}&A_{12}\\A_{21}&A_{22}\end{array}\right] \left[\begin{array}{c}x_1\\x_2\end{array}\right]+\left[\begin{array}{c}B_{1}\\B_{2}\end{array}\right]u+ \left[\begin{array}{c}D_{1}\\D_{2}\end{array}\right]W,\\
y &= Cx+CKW.
\end{aligned}
\label{divsde}
\end{equation}
where $x_1$ are measured nodes and $x_2$ are hidden nodes, and  $K=[K_1;K_2]$, $D_1=A_{11}K_1+A_{12}K_2$ and $D_2=A_{21}K_1+A_{22}K_2$.
Applying Laplace transformations and manipulating model~\eqref{divsde} yields:
\begin{equation}
\begin{aligned}
Y &= F_yY + F_uU+F_wW,
\end{aligned}
\label{predsf}
\end{equation}
where
\begin{equation}
\begin{aligned}
F_y &= \frac{1}{s}\left[A_{11}+A_{12}(sI-A_{22})^{-1}A_{21}\right],\\
F_u &= \frac{1}{s}\left[A_{12}(sI-A_{22})^{-1}B_{2}+B_{1}\right],\\
F_w &= \frac{1}{s}\left[A_{12}+A_{12}(sI-A_{22})^{-1}A_{22}\right]K_{2}+K_{1}.
\end{aligned}
\label{para_model}
\end{equation}
Here, $F_y$, $F_u$ and $F_w$ are transfer matrices, and each element is a proper transfer function, indicating the network is causal~\cite{yuan1}. Matrix $F_y$ represents the connectivity among measured nodes while matrices $F_u$ and $F_w$ relate inputs and noise to nodes, respectively. The topology of the network (i.e. model structure) is reflected by the zero structure of matrix $F_y$. For example, if $[F_y]_{rj}$ is zero, the $j$th node does not control the $r$th node. Note that internal dynamics of the network are described by transfer functions. The order of transfer functions is related to the number of hidden nodes, which is assumed to be unknown.

\subsection{Network identifiability}
The input-output map of model~\eqref{predsf} is as follows:
\begin{equation}
\begin{aligned}
Y = G_uU +G_wW,
\end{aligned}
\label{DSF}
\end{equation}
where $G=[G_u, G_w]$ and
\begin{equation}
\begin{aligned}
G_u=(I-F_y)^{-1}F_u,~
G_w=(I-F_y)^{-1}F_w.
\end{aligned}
\label{identifiability}
\end{equation}
For a network to be identifiable, matrices $F_y$, $F_u$ and $F_w$ must be uniquely inferred from equations~\eqref{identifiability}. A sufficient condition for identifiability is for matrix $F_w$ to be diagonal~\cite{qpmodel}. In what follows, assume $K_1$ is diagonal and $K_2$ is zero so that matrix $F_w$ equates to $K_1$. This means that Wiener processes affect directly only the measured nodes. Hence, given an input-output map $G$, transfer matrices $F_y$ and $F_w$ can be solved from the second equation in~\eqref{identifiability}: first, $F_w=K_1 = [G_w]_{s\rightarrow \infty}$, and then it follows that $(I-F_y)^{-1}K_1=G_w$ or $F_y=I-K_1G_w^{-1}$. Finally, $F_u=(I-F_y)G_u$ from the first equation in~\eqref{identifiability}. 
	
The target network we consider is stable and sparse. Hence, system matrix $A$ is assumed to be Hurwitz. As a result, $sF_y$ in~\eqref{para_model} is composed of stable transfer functions. In addition, $F_y$ is sparse. The objective is to identify matrices $F_y$, $F_u$ and $F_w$. 

\begin{rem}
	In the deterministic case, system matrices $(F_y,F_u)$ in our framework are related to the standard definition of DSFs~\cite{qpmodel}, $(W,V)$, as follows: $W=sF_y$ and $V=sF_u$.
\end{rem}

\subsection{The likelihood distribution}
To simplify notation, consider $u=0$, i.e. no deterministic input signals. The framework, however, can be easily extended to models with deterministic inputs. Note that each element of $F_y$ may have a pole at $0$ that dominates the dynamics of transfer functions. To avoid this problem, model~\eqref{predsf} is rewritten as
\begin{equation}
Y =\hat{F}_y\hat{Y}+W,
\label{dsf}
\end{equation}
where $\hat{F}_y= \frac{s}{s+a}F_y$, $\hat{Y}=\frac{s+a}{s}Y$ and $a$ is a positive constant. As a result, $\hat{F}_y$ is a stable and strictly proper transfer matrix. For each row $r$, $W_r(t)-W_r(s)=\mathcal{N}(0,\sigma_r(t-s))$, where $\sqrt{\sigma_r}$ is the $r$th diagonal element of $K_1$. 

To identify model~\eqref{dsf}, transfer functions are expressed non-parametrically using impulse responses. This approach avoids estimating model complexity and, more importantly,  effectively  promotes system stability.

 Assume $M$ time series of nodes are measured with low sampling frequencies at $T_1=\{t_{k_1},...,t_{k_q},...,t_{k_M}\}$ where $t_{k_1}=0$ and $k_{q+1}-k_q=N_q$ are positive integers.  Note that the sampling rate $t_{k_{q+1}}-t_{k_q}$ may not be a constant. In addition, all measurements are contaminated by white Gaussian noise. To improve inference accuracy, it is necessary to refine measured data at finer time instances and filter out measurement noise. Consider a series of finer time instances $T=\{t_0,...,t_i,...,t_N\}\supseteq T_1$ where $t_0=t_{k_1}$, $t_N=t_{k_M}$ and $t_i-t_{i-1} =\varDelta T_i$. Let $T_2=T\setminus T_1$ denote additional time instances apart from measurements. According to model~\eqref{dsf}, the dynamical system for the $r$th target node is:
\begin{equation}
\begin{aligned}
y_r(t_i)&= \sum_{j=1}^p\int_{0}^{t_i} f_{rj}(t_i-\tau)\hat{y}_j(\tau)d\tau +W_r(t_i),\ t_i\in T,\\
z_r(q) &= y_r(t_{k_q}) + e_r(q),
\label{dsf1}
\end{aligned}
\end{equation}
where $z(q)\in \mathcal{R}^p$ are noisy measurements of the network, $e(q)\in \mathcal{R}^p$ are i.i.d. white Gaussian noise with variance $\lambda$, $f_{rj}(\cdot)$ are impulse responses corresponding to transfer functions $[\hat{F}_y]_{rj}$, and $\hat{y}(t_i)\in  \mathcal{R}^p$ are the inverse Laplace transforms of $\hat{Y}$, i.e. $\hat{y}_j(t)=y_j(t)+a\int_0^ty_j(\tau)d\tau$.
 \begin{rem} The time increment $\varDelta T_i = t_i-t_{i-1}$ of finer time instances $ T$ may or may not be a constant, depending on system dynamics.  For example, finer sampling rates may be slowed down once initial fast dynamics wear off.
Nevertheless, for simplicity of exposition of the technique, this paper assumes a uniform time increment (i.e. $\varDelta T_i=\varDelta T$ for any index $i$). Our framework can easily be extended to cases of non-uniform time increments with minor modifications.
 \end{rem}
 
With noisy measurements collected with low sampling frequencies, the convolution integrals in~\eqref{dsf1} cannot be evaluated precisely. To achieve an accurate estimation, the integrals are discretized at finer time instances $T$.  Define the function $I(\cdot)$ as:
\begin{equation}
\begin{aligned}
I_{rj}(t_i)&= \sum_{v=0}^{i-1} f_{rj}(t_i-t_v)\hat{y}_j(t_v)(t_{v+1}-t_v),
\end{aligned}
\end{equation}    
where $t_i\in T$ and $\{t_v \}\subset T$.
Since $\lim_{\varDelta T\rightarrow 0}I_{rj}(t_i)= \int_{0}^{t_i} f_{rj}(t_i-\tau)\hat{y}_j(\tau)d\tau$, $I_{rj}(\cdot)$ can be a reasonable finite approximation of the integral if $\varDelta T$ is sufficiently small. Additionally, since stable impulse responses decay exponentially fast, they can be truncated when evaluating the convolution integrals. Assuming $|f_{rj}(k\varDelta T)|\approx 0$ for $k>l$, function $I(\cdot)$ is further approximated by:
\begin{equation}
\begin{aligned}
I_{rj}(t_i) &\approx \sum_{v=max\{0,i-l\}}^{i-1}f_{rj}(t_i-t_v)\hat{y}_j(t_v)(t_{v+1}-t_v).
\end{aligned}
\label{int}
\end{equation}     
\begin{rem} Given long time series and fast system dynamics, truncating impulse responses helps reduce computational costs of implementation without loosing too much accuracy. However, this approximation is not appropriate if data sources are limited and system dynamics are slow. Hence, section~\ref{MCMC} discusses a more complex approach to approximate impulse responses, which can deal with general cases.
\end{rem}

According to~\eqref{int}, model~\eqref{dsf1} evaluated at finer time instances is expressed in a linear regression form as follows (note that $l$ is set to $N$ if impulse responses are not truncated):
\begin{equation}
\begin{aligned}
y_r(t_i)&=\varDelta T\hat{Y}^{t_i}w_{r}+W_r(t_i),
\end{aligned}
\label{lik}
\end{equation}     
where 
\begin{equation}
\begin{aligned}
w_r &= [w_{r1};\cdots;w_{rp}],~w_{rj}=[f_{rj}(\varDelta T),\cdots,f_{rj}(l\varDelta T)]',\\
\hat{Y}^{t_i}&=[\hat{Y}_1^{t_i},\cdots,\hat{Y}_p^{t_i}],\\ \hat{Y}_j^{t_i}&=\left\{\begin{array}{ll}\left[\hat{y}_j(t_{i-1}),\cdots,\hat{y}_j(t_{i-l})\right]&if\  i\geqslant l \\
\left[\hat{y}_j(t_{i-1}),\cdots,\hat{y}_j(t_0),zeros(1,l-i)\right]&if\ i<l \end{array} \right.,\\ 
\hat{y}_j(t_i)&= y_j(t_i)+a\varDelta T\sum_{v=0}^{i-1}y_j(t_v).\\
\end{aligned}
\end{equation}
 Based on Bayes' rules, the likelihood distribution of model~\eqref{lik} for the refined trajectories is:
\begin{equation}
\begin{aligned}
&p(Y_{T_1},Y_{T_2}\big|w,\sigma)\\
&=\prod_{r=1}^p(2\pi\sigma_r\varDelta T)^{-\frac{N}{2}}\exp\left[-\frac{1}{2\sigma_r\varDelta T}\|\varDelta Y_r-\varDelta T \Phi w_r\|_2^2\right], 
\end{aligned}
\label{like}
\end{equation}
where $Y_{T_1}$ and $Y_{T_2}$ contain the refined trajectories at $T_1$ and $T_2$ time instances, respectively:
\begin{equation} 
\begin{aligned}
Y_{T_1} &= \left[y(t_{k_1}),\cdots,y(t_{k_M})\right]',\\
Y_{T_2} &= \left[y(t_{k_1+1}),\cdots,y(t_{k_2-1}),\cdots,y(t_{k_M-1})\right]',\\
\overline{Y}_r &= \left[y_r(t_N),\cdots, y_r(t_1)\right]',\\
\underline{Y}_r &= \left[y_r(t_{N-1}), \cdots, y_r(t_0)\right]',\\
\overline{\Phi} &= \left[\hat{Y}^{t_N};\cdots; \hat{Y}^{t_1}\right],\ \underline{\Phi} = \left[\hat{Y}^{t_{N-1}};\cdots ;\hat{Y}^{t_0}\right],\\
\varDelta Y_r&=\overline{Y}_r-\underline{Y}_r,
 \Phi=\overline{\Phi}-\underline{\Phi}.
\end{aligned}
\label{para}
\end{equation}
In addition, the likelihood distribution of noisy measurements according to~\eqref{dsf1} is the Gaussian distribution:
\begin{equation} 
\begin{aligned}
p(Z\big|Y_{T_1}) = \prod_{r=1}^p\prod_{q=1}^M\mathcal{N}(z_r(q)|y_r(t_{k_q}),\lambda),
\end{aligned}
\end{equation}
where $Z=[z(1),\cdots,z(M)]'$ contains measurements of all nodes and $\lambda$ is the measurement noise variance.

\subsection{The prior distributions}
To construct a full Bayesian model, prior distributions are introduced for all random quantities to capture prior knowledge~\cite{mur}. The major random quantities under consideration include refined trajectories, impulse responses, network topologies, and measurement and process noise variance. Additionally, hyperpriors are also introduced for the hyperparameters of prior distributions to complete the hierarchy.

To begin with, impulse responses are assumed to be independent Gaussian processes~\cite{gauss}. Gaussian processes are related to the theory of reproducing kernel Hilbert space (RKHS): covariance functions are equivalent to kernels of RKHS that control the property of functions in space~\cite{spline,nonp2}.  Whilst impulse responses are functions in kernel machine~\cite{spline,nonp1}, they are treated as stochastic processes under the Bayesian paradigm. To characterize stable impulse responses, three types of kernels are applied in our framework, including tuned/correlated kernels (TC), diagonal/correlated kernels (DC) and second order stable spline kernels (SS). These kernels have been widely used in kernel-based system identification to promote system stability~\cite{nonp1}. As a result, priors of impulse responses after truncation are independent Gaussian distributions:
\begin{equation}
\begin{aligned}
p(w|\gamma,\beta,s) = \prod_{r=1}^p\prod_{j=1}^{p}  \mathcal{N}(w_{rj}|0,K_{rj}),\\
\end{aligned}
\label{priw}
\end{equation} 
where $K_{rj}$ is the covariance matrix constructed from kernels, $\beta$ are hyperparameters of kernels, which control the decaying rate of impulse responses, $\gamma$ are scale variables of kernels, and $s$ are binary indicator variables implying network topology. If $s_{rj}=0$, impulse responses $w_{rj}$ are enforced to zero, meaning there is no link from node $j$ to node $r$.
\begin{equation}
\begin{aligned}
\left[K_{rj}\right]_{pq} &= s_{rj}|\gamma_{rj}|k(t_p,t_q;\beta_{rj}),\\
k_{TC}(t,s;\beta) &= \begin{array}{cc}\beta^{\max(t,s)},&\beta\in(0,1),\end{array}\\
k_{DC}(t,s;\beta) &= \begin{array}{cc}\beta_1^{\frac{(t+s)}{2}}\beta_2^{|t-s|},& \beta_1,\beta_2\in(0,1),\end{array}\\
k_{SS}(t,s;\beta) &= \begin{array}{cc}\frac{\beta^{t+s+\max(t,s)}}{2}-\frac{\beta^{3\max(t,s)}}{6},& \beta\in(0,1).\end{array}
\end{aligned}
\end{equation} 

The Inverse-Gamma distributions are assigned to both noise variance $\sigma$ and $\lambda$ as their conjugate priors. Without specific preference on their values, parameters $a_0$ and $b_0$ of the distributions are set to $0.001$, leading to non-informative priors. The probability density function of the Inverse-Gamma distribution is:
\begin{align}
p(x|a_0,b_0) &= IG(x;a_0,b_0)=\frac{b_0^{a_0}}{\Gamma(a_0)}x^{-a_0-1}e^{-\frac{b_0}{x}},
\end{align}
where $\Gamma(\cdot)$ is the gamma function.

Finally, hyperpriors are assigned to hyperparameters to complete the hierarchy. For hyperparameter $\gamma_{rj}$, the symmetric exponential distribution is applied as its prior: $p(\gamma_{rj};a_1)= \frac{a_1}{2}e^{-a_1|\gamma_{rj}|}$. For hyperparameter $\beta_{rj}$, the uniform distribution is employed:
\begin{equation}
\begin{aligned}
TC/SS:\ &p(\beta_{rj}) = 1,~\beta_{rj}\in(0,1),\\
DC:\ &p(\beta_{rj}) = 1,~ \beta_{rj1},\beta_{rj2}\in(0,1).
\end{aligned}
\end{equation}  
Assuming the probability of each true link is $p_s$, the prior distribution of topologies reflected by indicator variables $s$ is $p(s)=p_s^{N_s}(1-p_s)^{p^2-N_s}$ where $N_s=\sum_{r=1}^p\sum_{j=1}^ps_{rj}$ is the number of true links in the network.

\subsection{The full Bayesian model}
By incorporating the likelihood and prior distributions, the joint posterior distribution for model~\eqref{dsf1} is:
\begin{equation}
\begin{aligned}
&p(Y_{T_1},Y_{T_2},w,\sigma,\lambda,\beta,\gamma,s|Z)\\ 
&\propto p(Z|Y_{T_1},\lambda)p(Y_{T_1},Y_{T_2}|w,\sigma)p(w|\gamma,\beta,s)p(\sigma,\lambda,\beta,\gamma,s)\\
&\propto  \prod_{q=1}^M(2\pi\lambda)^{-\frac{p}{2}}\exp\left[-\frac{1}{2\lambda}\|z(q)-y(t_{k_q})\|_2^2\right]\\
&\times \prod_{r=1}^p\left\{|\Sigma_r|^{-\frac{1}{2}}\exp\left[-\frac{1}{2}(w_r-\mu_r)'\Sigma_r^{-1}(w_r-\mu_r)\right]\right.\\
&\times \left|\sigma_r\varDelta TI+\varDelta T^2 \Phi K_r \Phi'\right|^{-\frac{1}{2}}\\
&\left.\times \exp\left[-\frac{1}{2}\varDelta Y_r'(\sigma_r\varDelta TI+\varDelta T^2 \Phi K_r \Phi')^{-1}\varDelta Y_r\right]\right.\\
&\times \left. \lambda^{-a_0-1}\exp\left[-\frac{b_0}{\lambda}\right]\sigma_r^{-a_0-1}\exp\left[-\frac{b_0}{\sigma_r}\right]\right\}\\
&\times \left\{\prod_{r=1}^p\prod_{j=1}^p\exp\left[-a_1|\gamma_{rj}|\right]\right\}p_s^{N_s}(1-p_s)^{p^2-N_s},\\
\end{aligned}
\label{fullB}
\end{equation} 
where
\begin{equation}
\begin{aligned}
K_r&=blkdiag\{K_{r1},\cdots,K_{rp}\},\\
\Sigma_r^{-1} &= \frac{\varDelta T}{\sigma_r} \Phi'\Phi+K_r^{-1}, ~
\mu_r = \frac{1}{\sigma_r}\Sigma_r \Phi'\varDelta  Y_r.
\end{aligned}
\end{equation}
According to the full Bayesian model, the conditional posterior distributions of $w$, $\sigma$ and $\lambda$ have closed forms whilst those of the other random variables are only known up to a constant. This property determines how they are sampled in the following discussion.
\begin{equation}
\begin{aligned}
p(w|\beta,\gamma,s,\sigma,\lambda,Y_{T_1},Y_{T_2},Z)&=\prod_{r=1}^p\mathcal{N}(w_r|\mu_r,\Sigma_r),\\
p(\sigma|w,\beta,\gamma,s,\lambda,Y_{T_1},Y_{T_2},Z)&=\prod_{r=1}^pIG(\sigma_r;a_{\sigma_r},b_{\sigma_r}),\\
p(\lambda|w,\beta,\gamma,s,\sigma,Y_{T_1},Y_{T_2},Z)&=IG(\lambda;a_{\lambda},b_{\lambda}),\\
\end{aligned}
\label{pd}
\end{equation}  
 where
\begin{equation}
\begin{aligned}
a_{\lambda}&=a_0+\frac{pM}{2},b_{\lambda}=b_0+\frac{1}{2}\sum_{q=1}^M\|z(q)-y(t_{k_q})\|_2^2,\\
a_{\sigma_r}&=a_0+\frac{N}{2},~
b_{\sigma_r}=b_0+\frac{\|\varDelta Y_r-\varDelta T \Phi w_r\|_2^2}{2\varDelta T}.
\end{aligned}
\end{equation}  
By integrating out impulse responses $w$ from the full Bayesian model, one can deduce the marginal posterior distribution of refined trajectories and hyperparameters:
\begin{equation}
\begin{aligned}
&p(Y_{T_1},Y_{T_2},\beta,\gamma,s|\sigma,\lambda,Z)\\ 
&\propto  \prod_{q=1}^M(2\pi\lambda)^{-\frac{p}{2}}\exp\left[-\frac{1}{2\lambda}\|z(q)-y(t_{k_q})\|_2^2\right]\\
&\times \prod_{r=1}^p\left\{ \left|\sigma_r\varDelta TI+\varDelta T^2\Phi K_r \Phi'\right|^{-\frac{1}{2}}\right.\\
&\left.\times \exp\left[-\frac{1}{2}\varDelta Y_r'(\sigma_r\varDelta TI+\varDelta T^2\Phi K_r \Phi')^{-1}\varDelta Y_r\right]\right\}\\
&\times \left\{\prod_{r=1}^p\prod_{j=1}^p\exp\left[-a_1|\gamma_{rj}|\right]\right\}p_s^{N_s}(1-p_s)^{p^2-N_s}.
\end{aligned}
\label{marginal}
\end{equation}

\section{Network Inference using MCMC}\label{MCMC}
\subsection{The design of samplers}
With the full Bayesian model, one can estimate network topology, impulse responses and refined trajectories from the marginal posterior distributions $p(s|Z)$, $p(w|Z)$ and $p(Y_{T_1},Y_{T_2}|Z)$. However, this requires high-dimensional integrations, which are intractable for highly nonlinear distributions~\eqref{fullB}. Instead, approximate inference techniques can be applied to evaluate the distributions of interest~\cite{pattern}. This paper employs stochastic approximations based on numerical sampling to explore the full Bayesian model. In particular, Gibbs sampling and Metropolis-Hastings method (MH) are combined to draw samples from the target distribution (MH-within-Gibbs sampler~\cite{MHPCG}). Moreover, random variables $w$ are marginalized out from distributions to improve convergence properties~\cite{PCG}. The resulting sampler is a MH-within-PCG sampler.
 
 To begin with, the blocked Gibbs sampler~\cite{pattern} is proposed to draw samples from~\eqref{fullB} (Sampler 1). Random variables are sampled in sequence iteratively. 
 
\begin{center}
	\resizebox{\columnwidth}{!}{
	\begin{tabular}{l}
		\hline
		\textbf{Sampler 1}: Blocked Gibbs Sampler\\ 
		\hline
		1: Sample $p(Y_{T_1}^{k+1},Y_{T_2}^{k+1}|w^k,\beta^k,\gamma^k,s^k,\lambda^k,\sigma^k,Z)$\\
		2: Sample $p(\beta^{k+1},\gamma^{k+1},s^{k+1}|Y_{T_1}^{k+1},Y_{T_2}^{k+1},w^k,\sigma^k,\lambda^k,Z)$\\
		3: Sample $p(w^{k+1}|Y_{T_1}^{k+1},Y_{T_2}^{k+1},\beta^{k+1},\gamma^{k+1},s^{k+1},\sigma^k,\lambda^k,Z)$\\
		4: Sample $p(\sigma^{k+1}|Y_{T_1}^{k+1},Y_{T_2}^{k+1},w^{k+1},\beta^{k+1},\gamma^{k+1},s^{k+1},\lambda^k,Z)$\\
		5: Sample $p(\lambda^{k+1}|Y_{T_1}^{k+1},Y_{T_2}^{k+1},w^{k+1},\beta^{k+1},\gamma^{k+1},s^{k+1},\sigma^{k+1},Z)$ \\
		\hline
	\end{tabular}}
	\label{S1}
\end{center}

Sampler 1 requires full knowledge of conditional distributions of all random variables. Distributions of steps 3, 4 and 5 are presented in~\eqref{pd}, and can be sampled directly. However, the distributions of steps 1 and 2 are only known up to a constant. Hence, these two steps are conducted using MH methods, leading to a MH-within-Gibbs sampler. Moreover, to improve convergence properties, impulse responses $w$ are marginalized out from steps 1 and 2. It should be noticed that marginalization cannot be performed trivially. It has been shown in~\cite{MHPCG,PCG} that random variables to be marginalized out must be sampled immediately following the reduced MH steps. The resulting MH-within-Gibbs sampler is shown in Sampler 2.
\begin{center}
	\resizebox{\columnwidth}{!}{
	\begin{tabular}{l}
		\hline
		\textbf{Sampler 2}: MH-within-Gibbs\\ 
		\hline
		1: Sample $p(Y_{T_1}^{k+1},Y_{T_2}^{k+1}|\beta^k,\gamma^k,s^k,\lambda^k,\sigma^k,Z)$ (MH)\\
		2: Sample $p(w^{k+1}|Y_{T_1}^{k+1},Y_{T_2}^{k+1},\beta^k,\gamma^k,s^k,\lambda^k,\sigma^k,Z)$\\
		3: Sample $p(\beta^{k+1},\gamma^{k+1},s^{k+1}|Y_{T_1}^{k+1},Y_{T_2}^{k+1},\sigma^k,\lambda^k,Z)$ (MH)\\
		4: Sample $p(w^{k+1}|Y_{T_1}^{k+1},Y_{T_2}^{k+1},\beta^{k+1},\gamma^{k+1},s^{k+1},\lambda^k,\sigma^k,Z)$\\
		5: Sample $p(w^{k+1}|Y_{T_1}^{k+1},Y_{T_2}^{k+1},\beta^{k+1},\gamma^{k+1},s^{k+1},\sigma^k,\lambda^k,Z)$\\
		6: Sample $p(\sigma^{k+1}|Y_{T_1}^{k+1},Y_{T_2}^{k+1},w^{k+1},\beta^{k+1},\gamma^{k+1},s^{k+1},\lambda^k,Z)$\\
		7: Sample $p(\lambda^{k+1}|Y_{T_1}^{k+1},Y_{T_2}^{k+1},w^{k+1},\beta^{k+1},\gamma^{k+1},s^{k+1},\sigma^{k+1},Z)$ \\
		\hline
	\end{tabular}}
	\label{S2}
\end{center}

In Sampler 2, the sampled variables of steps 2 and 4 are not conditioned on  the next steps. Hence, these two steps can be removed from the sampler without affecting the invariant distribution of the Markov chain~\cite{MHPCG}. The final MH-within-PCG sampler is shown in Sampler 3. Step 1 samples the trajectories of nodes at finer time instances based on estimated system dynamics (sampled hyperparameters), which refines noisy time series that are sparsely and non-uniformly collected. Refined trajectories, in turn, correct the exploration of network topology and system dynamics in the following steps.
\begin{center}
	\resizebox{\columnwidth}{!}{
	\begin{tabular}{l}
		\hline
		\textbf{Sampler 3}: MH-within-PCG\\ 
		\hline
		1: Sample $p(Y_{T_1}^{k+1},Y_{T_2}^{k+1}|\beta^k,\gamma^k,s^k,\lambda^k,\sigma^k,Z)$ (MH)\\
		2: Sample $p(\beta^{k+1},\gamma^{k+1},s^{k+1}|Y_{T_1}^{k+1},Y_{T_2}^{k+1},\sigma^k,\lambda^k,Z)$ (MH)\\
		3: Sample $p(w^{k+1}|Y_{T_1}^{k+1},Y_{T_2}^{k+1},\beta^{k+1},\gamma^{k+1},s^{k+1},\sigma^k,\lambda^k,Z)$\\
		4: Sample $p(\sigma^{k+1}|Y_{T_1}^{k+1},Y_{T_2}^{k+1},w^{k+1},\beta^{k+1},\gamma^{k+1},s^{k+1},\lambda^k,Z)$\\
		5: Sample $p(\lambda^{k+1}|Y_{T_1}^{k+1},Y_{T_2}^{k+1},w^{k+1},\beta^{k+1},\gamma^{k+1},s^{k+1},\sigma^{k+1},Z)$ \\
		\hline
	\end{tabular}}
	\label{S3}
\end{center}

In Sampler 3, steps 3, 4 and 5 can be implemented directly whilst steps 1 and 2 employ MH algorithms. Next, we focus on designing steps 1 (trajectories sampling) and 2 (hyperparameters smapling).

\subsection{Sample trajectories}\label{tra0}
 The marginal posterior distribution of refined trajectories is as follows.
\begin{equation}
\begin{aligned}
&p(Y_{T_1},Y_{T_2}|\beta,\gamma,s,\sigma,\lambda,Z)\\ 
&\propto  \prod_{q=1}^M(2\pi\lambda)^{-\frac{p}{2}}\exp\left[-\frac{1}{2\lambda}\|z(q)-y(t_{k_q})\|_2^2\right]\\
&\times \prod_{r=1}^p\left\{ \left|\sigma_r\varDelta TI+\varDelta T^2 \Phi K_r \Phi'\right|^{-\frac{1}{2}}\right.\\
&\left.\times \exp\left[-\frac{1}{2}\varDelta Y_r'(\sigma_r\varDelta TI+\varDelta T^2\Phi K_r \Phi')^{-1}\varDelta Y_r\right]\right\}.\\
\end{aligned}
\label{tra1}
\end{equation}  
According to MH methods, random variables are drawn from proposal distributions in each iteration. The proposed random variables are either accepted or rejected depending on the acceptance probability. In order to produce reasonable proposals for refined trajectories, the proposal distribution is formed as $q(Y_{T_1}^p,Y_{T_2}^p|Y_{T_1}^k,Y_{T_2}^k)=q(Y_{T_2}^p|Y_{T_1}^p,Y_{T_2}^k)q(Y_{T_1}^p|Y_{T_1}^k)$. The refined trajectories at the measurement time instances ($Y_{T_1}$) are sampled first based on the current Markov state $Y_{T_1}^k$. Provided the proposed $Y_{T_1}^p$ and the current $Y_{T_2}^k$, the refined trajectories at finer time instances ($Y_{T_2}$) are sampled to connect each pair of adjacent trajectory points in $Y_{T_1}^p$. According to~\eqref{dsf1}, it is relatively easy to sample $Y_{T_1}$ since they are normally distributed given measurements $Z$, i.e. 
\begin{align}
p(Y_{T_1}|Z)=\prod_{r=1}^p\mathcal{N}(Y_{r,T_1}|Z_r,\lambda),
\end{align}
where 
\begin{equation}
\begin{aligned}
Z_r &= Z(:,r) = [z_r(1),...,z_r(M)]',\\
Y_{r,T_1} &= Y_{T_1}(:,r) = [y_r(t_{k_1}),...,y_r(t_{k_M})]'.\\
\end{aligned}
\end{equation}   
 However, sampling $Y_{T_2}$ is much more complex. The target distribution~\eqref{tra1} is highly nonlinear with respect to $Y_{T_2}$. To achieve reasonable acceptance probability, the proposal distribution should be somehow consistent with the target distribution.

Random walk sampling~\cite{pattern} is applied to generate proposals for trajectories. Random walk uses the Gaussian distribution to generate proposals, which is easy to implement. However, standard random walk suffers from the curse of dimensionality. As the length of refined trajectories increases, the acceptance probability of Markov chains may drop dramatically. If one improves acceptance probability by reducing the step size, the resulting algorithm may require much longer iterations to converge. Hence, it is essential to employ a sampling scheme that is robust to the dimension of sampling space. A possible solution is the preconditioned Crank-Nicolson sampler (pCN~\cite{pCN}). The proposal of pCN is of AR(1) type rather than a centred random walk. For the target distribution $p(x)\varpropto\mathcal{N}(x|m,\sigma)f(x)$ where $f(x)$ is some real-valued function, pCN generates the proposal as $x^{p}=\sqrt{1-\varepsilon^2}x^k+(1-\sqrt{1-\varepsilon^2})m+\varepsilon\mathcal{N}(0,\sigma)$ where $\varepsilon$ is the step size. When calculating the acceptance probability, the ratio of proposal distributions cancels out with that of the Gaussian parts of the target distribution. As a result, the acceptance probability only depends on the residual function $f(x)$ (i.e. $A(x^p|x^k)=\min\{1,\frac{f(x^p)}{f(x^k)}\}$). Given normally distributed target distributions, proposals of pCN are accepted with probability $1$, regardless of the dimension of random variables and the step size. In this case, pCN is as efficient as sampling the Gaussian distribution directly, which cannot be achieved by standard random walk.

By using the pCN sampling, the proposal for $Y_{T_1}$ is drawn from distribution $q(Y_{T_1}^p|Y_{T_1}^k)=\prod_{r=1}^p\mathcal{N}(Z_r+\sqrt{1-\varepsilon^2}(Y_{r,T_1}^k-Z_r),\varepsilon^2\lambda)$. Consequently, the Gaussian part of the target distribution~\eqref{tra1} related to measurements $Z$ cancels out with the proposal distribution in the acceptance probability.

Next is to design the proposal for $Y_{T_2}$ using pCN. Considering the  terms associated with $Y_{T_2}$ in~\eqref{tra1} are proportional to conditional distribution $p(Y_{T_2}|Y_{T_1},\beta,\gamma,s,\sigma,\lambda)$, we first decompose this distribution into Gaussian and non-Gaussian parts. By using the Woodbury formula, distribution $p(Y_{T_2}|Y_{T_1},\beta,\gamma,s,\sigma,\lambda)$ can be decomposed as:
\begin{equation}
\begin{aligned}
p(Y_{T_2}|Y_{T_1},\beta,\gamma,s,\sigma,\lambda)
\varpropto \prod_{r=1}^pp_{r,1}(Y_{T_2}|Y_{T_1})p_{r,2}(Y_{T_2}|Y_{T_1}),\\
\end{aligned}
\end{equation}   
where parameters independent on $Y_{T_2}$ are ignored for simplicity.
\begin{equation}{\tiny
\begin{aligned}
&p_{r,1}(Y_{T_2}|Y_{T_1})\varpropto
 exp\left[-\frac{1}{2\sigma_r\varDelta T}\|\varDelta Y_r\|_2^2\right],\\
&p_{r,2}(Y_{T_2}|Y_{T_1})\\
&\varpropto|\sigma_r K_r^{-1}+\varDelta T\Phi'\Phi|^{-\frac{1}{2}}exp\left[\frac{1}{2\sigma_r}\varDelta Y_r'\Phi(\sigma_r K_r^{-1}+\varDelta T\Phi'\Phi)^{-1}\Phi'\varDelta Y_r\right].\\
\end{aligned}}
\end{equation}   
Distribution $p_{r,1}(Y_{T_2}|Y_{T_1})$ can be cast into the Gaussian form since its logarithm is a quadratic function of $Y_{T_2}$:
\begin{equation}{\tiny
\begin{aligned}
&p_{r,1}(Y_{T_2}|Y_{T_1})\\
&
\varpropto \exp\left\{-\frac{1}{2\sigma_r\varDelta T}\sum_{j=1}^{M-1}[Y_{r,j,T_2}-m_{r,j}]'C_j^{-1}[Y_{r,j,T_2}-m_{r,j}]\right\}\\
&\times \exp\left\{-\frac{1}{2\sigma_r\varDelta T}\varDelta Y_{r,T_1}'\bar{N}^{-1}\varDelta Y_{r,T_1}\right\},\\
\end{aligned}}
\label{BB}
\end{equation}  
where $Y_{r,j,T_2}\in\mathcal{R}^{N_j-1}$, $m_{r,j}\in\mathcal{R}^{N_j-1}$, $C_{j}\in\mathcal{S_{+}}^{N_j\times N_j}$ and $\varDelta Y_{r,T_1}\in\mathcal{R}^{M-1}$.
\begin{equation}
 \begin{aligned}
 \varDelta Y_{r,T_1} &=[y_r(t_{k_M})-y_r(t_{k_{M-1}}),\cdots,y_r(t_{k_2})-y_r(t_{k_{1}})]',\\
Y_{r,j,T_2}&=[y_r(t_{k_j+1}),\cdots,y_r(t_{k_{j+1}-1})]',\\
[m_{r,j}]_i&=y_r(t_{k_j})+\frac{i}{N_j}[y_r(t_{k_{j+1}})-y_r(t_{k_j})],\\
[C_j]_{pq}&=\frac{p(N_j-q)}{N_j}, p\leq q,~
\bar{N} = diag\{N_{M-1},\cdots,N_1 \}.
 \end{aligned}
\end{equation}  
 According to~\eqref{BB}, $Y_{T_2}$ are partially described as Brownian bridges that connect adjacent points of $Y_{T_1}$. Following the pCN framework, the proposal distribution for $Y_{T_2}$ is designed based on $p_{r,1}(Y_{T_2}|Y_{T_1})$ while $p_{r,2}(Y_{T_2}|Y_{T_1})$ is regarded as the residual function.  Hence, the refined trajectories between measurement time instances $t_{k_j}$ and $t_{k_{j+1}}$ are proposed based on $Y_{T_1}^p$ as $Y_{r,j,T_2}^p= m_{r,j}^p+\sqrt{1-\varepsilon^2}(Y_{r,j,T_2}^k-m_{r,j}^k)+\varepsilon \mathcal{N}(0,\sigma_r\varDelta TC_j)$. Similar to $Y_{T_1}$, the Gaussian part of $Y_{T_2}$ in~\eqref{tra1} after decomposition cancels out with the proposal distribution when calculating the acceptance probability.

The resulting acceptance probability of trajectories is:
\begin{equation}{\tiny
\begin{aligned}
&A(Y_{T1}^{p},Y_{T2}^{p}|Y_{T1}^k,Y_{T2}^k)
=\min\{1,r(Y_{T1}^{p},Y_{T2}^{p}|Y_{T1}^k,Y_{T2}^k)\},\\
&r(Y_{T1}^{p},Y_{T2}^{p}|Y_{T1}^k,Y_{T2}^k)
=\frac{p(Y_{T1}^{p},Y_{T2}^{p})q(Y_{T1}^k|Y_{T1}^{p})q(Y_{T2}^k|Y_{T1}^{k},Y_{T2}^{p})}{p(Y_{T1}^k,Y_{T2}^k)q(Y_{T1}^p|Y_{T1}^{k})q(Y_{T2}^p|Y_{T1}^{p},Y_{T2}^{k})},
\end{aligned}}
\end{equation}
where $r(Y_{T1}^{p},Y_{T2}^{p}|Y_{T1}^k,Y_{T2}^k)$ is the acceptance ratio.
\begin{equation}{\tiny
	\begin{aligned}
	&r(Y_{T1}^{p},Y_{T2}^{p}|Y_{T1}^k,Y_{T2}^k)\\
	&=\prod_{r=1}^p\left\{\frac{|\sigma_rK^{-1}_r+\varDelta T (\Phi^p)'\Phi^p |^{-\frac{1}{2}}}{|\sigma_rK^{-1}_r+\varDelta T (\Phi^k)'\Phi^k |^{-\frac{1}{2}}}\frac{exp\left\{-\frac{1}{2\sigma_r\varDelta T}(\varDelta Y_{r,T_1}^p)'\bar{N}^{-1}\varDelta Y_{r,T_1}^p\right\}}{exp\left\{-\frac{1}{2\sigma_r\varDelta T}(\varDelta Y_{r,T_1}^k)'\bar{N}^{-1}\varDelta Y_{r,T_1}^k\right\}}\right.\\
	&\times \left.\frac{exp\{\frac{1}{2\sigma_r}(\varDelta Y^p_r)'\Phi^p(\sigma_r K_r^{-1}+\varDelta T(\Phi^p)'\Phi^p)^{-1}(\Phi^p)'\varDelta Y_r^p \}}{exp\{\frac{1}{2\sigma_r}(\varDelta Y^k_r)'\Phi^k(\sigma_r K_r^{-1}+\varDelta T(\Phi^k)'\Phi^k)^{-1}(\Phi^k)'\varDelta Y_r^k \}}\right\}.\\
	\end{aligned}}
\end{equation}

\subsection{Sample hyperparameters}
According to~\eqref{marginal}, each group of $(\beta,\gamma,s)$ is conditionally independent. Hence, they can be sampled in parallel. While sampling hyperparameters $\beta$ and $\gamma$ explores internal dynamics, sampling indicator variables $s$ is crucial for topology detection. To traverse across different topologies effectively, two types of moves are proposed for indicator variables $s$. In each iteration, only one move is implemented to update random variables. 

\textbf{\textit{Switch Move}:}
Switch move is executed with probability $P_S$. The topology of the next state only differs from the current state in one link. A link is either added or removed from the network depending on the current state. For example, if $s_r^k$ is $[\begin{array}{ccc}1& 0& 1\end{array}]$,  possible proposals for $s_r^{k+1}$ include $[\begin{array}{ccc}0& 0& 1\end{array}]$, $[\begin{array}{ccc}1& 1& 1\end{array}]$ and $[\begin{array}{ccc}1& 0& 0\end{array}]$. If $s_{rj}^{k+1}-s_{rj}^k=\pm1$, link $j\rightarrow r$ is added/removed to the current topology. Only the hyperparameters corresponding to the switched link are updated.

\textbf{\textit{Update Move}:}
Update move is executed with probability $P_U$. The topology of the network is unchanged in the next state (i.e. $s_r^{k+1}=s_r^k$) whilst the hyperparameters of the true links of the current topology are updated.

As a result, the proposal distribution for the $r$th target node is:
\begin{equation}
\begin{aligned}
&q_m(\beta_r^p,\gamma_r^p,s_r^p|\beta_r^k,\gamma_r^k,s_r^k)\\
&=P_mq_m(s_r^p|s_r^k)q_m(\beta_r^p|\beta_r^k)q_m(\lambda_r^p|\lambda_r^k),
\end{aligned}
\end{equation}
where subscript $m$ denotes move types, and $P_S=0.6$ and $P_U=0.4$ present the probability of executing switch and update moves, respectively.

The acceptance probability for proposals is:
\begin{equation}{
\begin{aligned}
&A_m(\beta_r^p,\gamma_r^p,s_r^p|\beta_r^k,\gamma_r^k,s_r^k)\\
&=\min\{1,r_m(\beta_r^p,\gamma_r^p,s_r^p|\beta_r^t,\gamma_r^t,s_r^t)\}\\
&=\min\left\{1,\frac{p(\beta_r^p,\gamma_r^p,s_r^p)q_m(\beta_r^k,\gamma_r^k,s_r^k|\beta_r^p,\gamma_r^p,s_r^p)}{p(\beta_r^k,\gamma_r^k,s_r^k)q_m(\beta_r^p,\gamma_r^p,s_r^p|\beta_r^k,\gamma_r^k,s_r^k)}\right\}.\\
\end{aligned}}
\end{equation}

\subsubsection{Switch Move}
To realize the switch move, the link to be modified is uniformly selected from all true and null links of the current topology. The corresponding indicator variable is switched between $1$ and $0$. In addition, hyperparameters of the switched links are sampled with other components unchanged.  Assuming link $j\rightarrow r$ is switched, the acceptance ratio of the switch move is:
\begin{equation}
\begin{aligned}
&r_S(\beta_r^p,\gamma_r^p,s_r^p|\beta_r^k,\gamma_r^k,s_r^k)\\
&=\left(\frac{p_s}{1-p_s}\right)^{s_{rj}^p-s_{rj}^k}\frac{|\sigma_r(K^{p}_r)^{-1}+\varDelta T \Phi'\Phi |^{-\frac{1}{2}}|K_{r}^p|^{-\frac{1}{2}}}{|\sigma_r(K^{k}_r)^{-1}+\varDelta T \Phi'\Phi |^{-\frac{1}{2}}|K_{r}^k|^{-\frac{1}{2}}}\\
&\times\frac{\exp\{\frac{1}{2\sigma_r}\varDelta Y_r'\Phi(\sigma_r (K_r^{p})^{-1}+\varDelta T\Phi'\Phi)^{-1}\Phi'\varDelta Y_r \}}{\exp\{\frac{1}{2\sigma_r}\varDelta Y_r'\Phi(\sigma_r (K_r^{k})^{-1}+\varDelta T\Phi'\Phi)^{-1}\Phi'\varDelta Y_r \}}\\
&\times \exp[-a_1(|\gamma_{rj}^p|-|\gamma_{rj}^k|)].\\
\end{aligned}
\end{equation} 

The procedure of the switch move is summarized in Algorithm~\ref{S}.
\begin{algorithm}[!]
	\caption{Switch Move}
	\label{S}
	\begin{algorithmic}[1]
		\State With probability $P_S$, choose Switch Move.
		\State Select number $j$ uniformly from $\{1,\cdots,p\}$ and switch the value of $s_{rj}$ (i.e. $s_{rj}^p=1-s_{rj}^k$).	
		\State Draw proposals $\beta^p_{rj}$ and $\gamma^p_{rj}$ from $q_S(\beta_{rj})$ and $q_S(\gamma_{rj})$ as follows whilst keep the other components unchanged.
		\begin{equation}
		\begin{aligned}
		q_S(\gamma_{rj})&=\mathcal{N}(\gamma_{rj}|\gamma_{rj}^k,\varepsilon^2)\\
		TC/SS:\ q_S(\beta_{rj}) &= U(\beta_{rj};0,1)\\
		DC:\ q_S(\beta_{rj}) &= U(\beta_{rj1};0,1)U(\beta_{rj2};0,1)\\
		\end{aligned}
		\end{equation}
		\State Accept with probability $A_S$.
	\end{algorithmic}
\end{algorithm}

\subsubsection{Update Move}

With the update move, only the hyperparameters corresponding to true links are sampled where each component is drawn from proposal distributions independently. Since the hyperparameters of null links do not contribute to system dynamics, the acceptance probability of these hyperparameters purely depends on their prior distributions. Hence, it is reasonable to exclude them from the update move. 

To begin with, hyperparameters $\gamma_r$ are sampled using the standard random walk. With step size $\varepsilon$, the proposal for $\gamma_{rj}$ of true link $j\rightarrow r$ is generated as $\gamma_{rj}^p=\gamma_{rj}^k+\varepsilon\mathcal{N}(0,1)$.

For hyperparameters $\beta_r$, the proposals are drawn from the following distribution. For a random variable $\theta\in[l,u]$ with expected value $\bar{\theta}$, the probability density function is defined as:
\begin{equation}
\begin{aligned}
p_U(\theta;\bar{\theta},l,u,\varepsilon)=\left\{\begin{array}{ll}U(\bar{\theta}-\frac{\varepsilon}{2},\bar{\theta}+\frac{\varepsilon}{2}) &l+\frac{\varepsilon}{2}<\bar{\theta}< u-\frac{\varepsilon}{2}\\ U(l,l+\varepsilon)& \bar{\theta}\leq l+\frac{\varepsilon}{2}\\U(u-\varepsilon,u)& \bar{\theta}\geq u-\frac{\varepsilon}{2}\end{array},\right. 
\end{aligned}
\end{equation}
where $U(a,b)$ is the uniform distribution on $(a,b)$, and $\varepsilon$ is the selection window for sampling. Hence, the proposal distribution for $\beta_{rj}$ of true link $j\rightarrow r$ is:
\begin{equation}
\begin{aligned}
TC/SS: q_U(\beta_{rj}^p|\beta_{rj}^k)&= p_U(\beta_{rj}^p;\beta_{rj}^k,0,1,0.1),\\
DC:q_U(\beta_{rj}^p|\beta_{rj}^k)&=p_U(\beta_{rj1}^p;\beta_{rj1}^k,0,1,0.1)\\
&\times p_U(\beta_{rj2}^p;\beta_{rj2}^k,0,1,0.1).\\
\end{aligned}
\end{equation}
The acceptance ratio for the update move is:
\begin{equation}{
\begin{aligned}
&r_U(\beta_r^p,\lambda_r^p,s_r^p|\beta_r^k,\lambda_r^k,s_r^k)\\
&=\left(\prod_{j=1}^p\frac{exp\left\{-a_1(|\gamma_{rj}^p|) \right\}}{exp\left\{-a_1(|\gamma_{rj}^k|)\right\}}\right)\frac{|\sigma_r(K^{p}_r)^{-1}+\varDelta T \Phi'\Phi |^{-\frac{1}{2}}}{|\sigma_r(K^{k}_r)^{-1}+\varDelta T \Phi'\Phi |^{-\frac{1}{2}}}\\
&\times\frac{|K_{r}^p|^{-\frac{1}{2}}\exp\{\frac{1}{2\sigma_r}\varDelta Y_r'\Phi(\sigma_r (K_r^{p})^{-1}+\varDelta T\Phi'\Phi)^{-1}\Phi'\varDelta Y_r \}}{|K_{r}^k|^{-\frac{1}{2}}\exp\{\frac{1}{2\sigma_r}\varDelta Y_r'\Phi(\sigma_r (K_r^{k})^{-1}+\varDelta T\Phi'\Phi)^{-1}\Phi'\varDelta Y_r \}}.\\
\end{aligned}}
\end{equation} 

The procedure of the update move is summarized in Algorithm~\ref{U}.
\begin{algorithm}[!]
	\caption{Update Move}
	\label{U}
	\begin{algorithmic}[1]
		\State With probability $P_U$, choose Update Move.
		\State Draw proposals $\beta^p_r$ and $\gamma^p_r$ from $q_U(\beta_r^p|\beta_r^k)$ and $q_U(\gamma_r^p|\gamma_r^k)$, respectively.
		\State Accept with probability $A_U$.
	\end{algorithmic}
\end{algorithm}

\subsection{MCMC sampler for the full Bayesian model}
To conclude, Algorithm~\ref{final} presents the sampling procedure of the full Bayesian model.

\begin{algorithm}[!]
	\caption{MCMC sampler}
	\label{final}
	\begin{algorithmic}[1]
		\State Initialize $w^0$, $\beta^0$, $\gamma^0$, $\lambda^0$, $\sigma^0$, $Y^0$ and $s^0$.
		\For  {$k=1:k_{max}$}
		\State Sample refined trajectories ($Y_{T_1}$ and $Y_{T_2}$) according to section~\ref{tra0}.
		\For {$r=1:p$}
		\State Sample $P_{m}$ from $U(0,1)$.
		\If {$P_{m}\leq P_S$}
		\State Execute Switch Move (Algorithm~\ref{S}).
		\Else
		\State Execute Update Move (Algorithm~\ref{U}).
		\EndIf
		\State Sample $w_r$ according to~\eqref{pd}.
		\State Sample $\sigma_r$ according to~\eqref{pd}.
		\EndFor
		\State Sample $\lambda$ according to~\eqref{pd}.
		\EndFor
		\State Store all the samples.
	\end{algorithmic}
\end{algorithm}

\subsection{Algorithm implementation}

Without truncating impulse responses (i.e. impulse reponses are evaluated at all finer time instances), one has to calculate the inversion and determinant of matrices $K_r^{-1}$  and $(\sigma_r K_r^{-1}+\varDelta T \Phi'\Phi)$ in each iteration, where standard operations are costly and require $O(p^3N^3)$ work. In addition, since the diagonal elements of matrix $K_r$ decay exponentially fast, $K_r$ becomes ill-conditioned as the number of finer time instances grows~\cite{Chen}. Hence, it is essential to reduce the matrix dimension given large datasets.

The main idea is to apply sparse likelihood approximation~\cite{sla,pseu}.  According to the previous formulation, impulse responses are evaluated at finer time instances $T=\{\varDelta T,...,N\varDelta T\}$. Let subset $D\subset T$ and $R=T\setminus D$. We use impulse responses at time instances $D$ ($w^D$) to approximate those at $R$ ($w^R$). Since the prior distribution of $w_{rj}$ is Gaussian ($\mathcal{N}(w_{rj}|0,K_{rj})$ referring to~\eqref{priw}), $w_{rj}^R$ conditioned on $w_{rj}^D$ is also normally distributed:
\begin{equation}{
\begin{aligned}
p(w_{rj}^R|w_{rj}^D)&=\mathcal{N}(\mu,\Sigma), \\
\end{aligned}}
\label{condi}
\end{equation}   
where
\begin{equation}{
	\begin{aligned}
	\mu&=K_{rj}^{RD}(K_{rj}^{DD})^{-1}w_{rj}^D,\\
	\Sigma&=K_{rj}^{RR}-K_{rj}^{RD}(K_{rj}^{DD})^{-1}K_{rj}^{DR},\\
	K_{rj}^{DD}&=E(w_{rj}^D(w^D_{rj})'),\\
	K_{rj}^{RR}&=E(w_{rj}^R(w_{rj}^R)'),\\
	K_{rj}^{RD}&=(K_{rj}^{DR})'=E(w_{rj}^R(w_{rj}^D)').\\
	\end{aligned}}
\end{equation}   
 Given $w_{rj}^D$, impulse responses $w_{rj}^R$ are approximated as the mean of conditional distribution~\eqref{condi}. Consequently, impulse responses $w_{rj}$ are described using $w_{rj}^D$ as:
 \begin{equation}{
\begin{aligned}
w_{rj}&=D_{rj}w_{rj}^D,\\
\end{aligned}}
\label{aw}
 \end{equation}   
 where $P$ is a permutation matrix and
\begin{equation}{
\begin{aligned}
D_{rj}=P\left[\begin{array}{c}I\\K_{rj}^{RD}(K_{rj}^{DD})^{-1}\end{array}\right].\\
\end{aligned}}
\end{equation} 
Based on~\eqref{aw}, the likelihood distribution of refined trajectories and the prior distribution of impulse responses become:
\begin{equation}{\tiny
\begin{aligned}
&p(Y_{T_1},Y_{T_2}|w^D,\gamma,\beta,s) \\
&=\prod_{r=1}^p(2\pi\sigma_r  \varDelta T)^{-\frac{N}{2}} \exp\left\{-\frac{1}{2\sigma_r  \varDelta T}\|\varDelta Y_r-\varDelta T  \Phi D_rw^D_r\|^2_2\right\},\\
&p(w^D|\gamma,\beta,s) = \prod_{r=1}^p\prod_{j=1}^{p}  \mathcal{N}(w_{rj}^D|0,K_{rj}^{DD}),\\
\end{aligned}}
\label{al}
\end{equation}   
where  $D_r=blkdiag\{D_{r1},...,D_{rp}\}$ and $K_r^{DD}=blkdiag\{K_{r1}^{DD},...,K_{rp}^{DD}\}$.

Substituting the distributions~\eqref{al} into the full Bayesian model~\eqref{fullB}, the major matrix to be tackled becomes $(\sigma_r (K_r^{DD})^{-1}+\varDelta TD_r'\Phi'\Phi D_r)$. Assuming the cardinality of $D$ is $d$, the computational cost of calculating matrix inversion and determinant reduces to $O(p^3d^3)$. Additionally, the conditional number of $K_r^{DD}$ is improved by selecting $D$ properly.
\begin{rem} Note that the time instances of the selected partial impulse responses $w^D$ are not necessarily equispaced. Since impulse responses decay exponentially fast, the major system dynamics are contained in early time instances. Therefore, it is reasonable to select more points before time $l\varDelta T$ (if $|f_{rj}(k\varDelta T)|\approx 0$ at $k>l$) but fewer points afterwards. By excluding impulse responses close to zero, the conditional number of $K_r^{DD}$ is also improved, which enhances numerical stability when calculating matrix inversion and determinant. 
\end{rem}
\begin{rem}
Impulse responses $w^D$ can be evaluated at pseudo-time points that are not included in the finer time instances (e.g. $D\nsubseteq T$ and $R=T$). The corresponding full Bayesian model is constructed following the same procedure discussed in this section.
\end{rem}

\subsection{Detection of network topology}
Detection of network topology is based on the marginal posterior distribution of indicator variables, $p(s|Z)$.  We discard the first half of samples (i.e. burn-in period) to ensure that the remaining samples follow the target distribution. Then,  the distribution can be estimated using its empirical counterpart constructed from samples:
\begin{equation}
\begin{aligned}
P(s|Z) &\approx  \frac{2}{k_{max}}\sum_{k=\frac{k_{max}}{2}}^{k_{max}} \mathbf{1}_{s}(s^k),
\end{aligned}
\label{emp}
\end{equation}
where
\begin{equation}
\begin{aligned}
\mathbf{1}_{x}(y) &= \left\{\begin{array}{cc}1&~if~y=x\\0&~if~y\neq x\end{array}\right..
\end{aligned}
\end{equation}
Given empirical distribution~\eqref{emp}, the most likely network topology is estimated using maximum a posteriori (MAP): $s_{opt}=\max_{s}P(s|Z)$. In addition, the confidence of all putative links can be evaluated via their probability distribution. For example, the probability of link $j\rightarrow r$ is estimated as:
\begin{equation}
\begin{aligned}
P(s_{rj}=1|Z) &\approx  \frac{2}{k_{max}}\sum_{k=\frac{k_{max}}{2}}^{k_{max}} s_{rj}^k.
\end{aligned}
\end{equation}
Finally, impulse responses and refined trajectories are estimated as follows:
\begin{equation}
\begin{aligned}
\hat{w} &=E(w|Z)= \frac{2}{k_{max}}\sum_{k=\frac{k_{max}}{2}}^{k_{max}}w^k, \\
\hat{Y} &=E(Y|Z)= \frac{2}{k_{max}}\sum_{k=\frac{k_{max}}{2}}^{k_{max}}Y^k.
\end{aligned}
\end{equation}

\section{SIMULATION}\label{Simulation}
We conducted two Monte Carlo simulations and compared our method with several state-of-the-art methods including GSBL, ARNI~\cite{ARNI}, dynGENIE3~\cite{dynGENIE3}, GENIE3~\cite{GENIE3}, kernel-based approaches~\cite{nonp} and BINGO~\cite{BINGO}. The first part of simulations generated random linear continuous-time networks with sparse topologies (including a ring structure). The networks were described by linear SDE models in~\eqref{sde} with additional inputs. The models were simulated under various noise levels and inferred using time series data of different lengths. The second set of simulations tested all algorithms on a synthetic biological network of yeast. Note that this artificially designed network produced real experimental data.

Four criteria are applied to evaluate the performance of algorithms including true positive rate (TPR), precision (PREC), area under the receiver operating characteristic curve (AUROC) and area under the precision recall curve (AUPREC)~\cite{Aderhold2014}. TPR shows the percentage of the true links in the ground truths that were successfully inferred. PREC equates to the rate of the correct links over the all inferred. TPR and PREC indicate the accuracy of the inferred network. Whilst TPR and PREC are applied to evaluate the most likely network, AUROC and AUPREC are more suitable for the methods that estimate the confidence of all links (e.g. dynGENIE3, GENIE3, ARNI and BINGO). AUROC and AUPREC reveal similar information of TPR and PREC.

\subsection{Random SDE networks}

$100$ networks were generated with random topologies and internal dynamics. All networks contained $15$ nodes in total with $10$ nodes measured. The networks were described by linear SDEs where nodes were represented by state variables. Sparse matrices $A\in\mathcal{R}^{15\times15}$ were first produced randomly using the function $sprandn(n,n,density)$ in Matlab. Only Hurwitz matrices $A$ (eigenvalues with negative real parts) were kept for simulation. Only the first $10$ states (nodes) were measured for inference and measurements were contaminated with white Gaussian noise. Each measured node was independently driven by an input (known) and process noise (unknown).

\begin{figure}
	\centering
	\includegraphics[width=0.6\columnwidth]{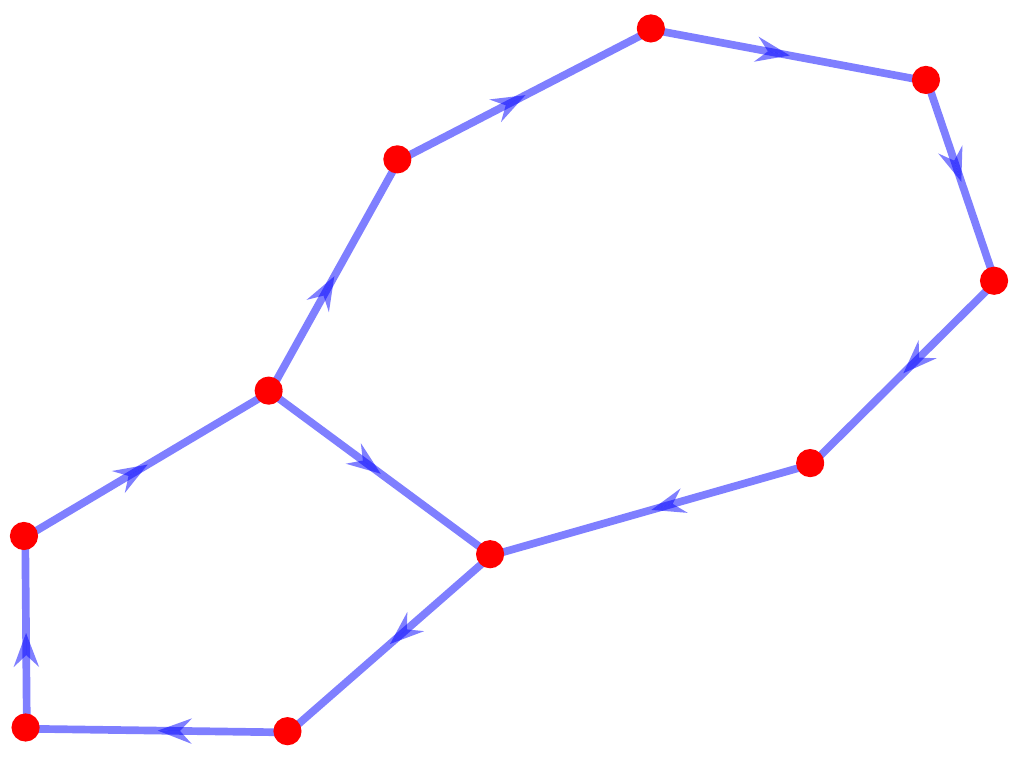}
	\caption{A random network of measured nodes. Solid lines with arrows represent links. Red circles denote nodes.}
	\label{random}
\end{figure}

To simulate the models, inputs and process noise were independent Wiener processes ($W(t)-W(s)=\mathcal{N}(0,\sigma(t-s))$) where $\sigma$ is the variance of Wiener processes. The variance of inputs was fixed to $1$ whilst that of process noise varied. The Signal-to-Noise ratio is defined as $SNR=10\log\frac{\sigma_u}{\sigma_e}$ where $\sigma_u$ and $\sigma_e$ are signal variance of inputs and noise, respectively. To allow a comparison with the methods that require equispaced sampling, time series data were collected with sampling frequency $1$ and various lengths between $40$ to $140$. The noise variance of measurement noise was set to $10^{-3}$. The increment step ($\varDelta T$) for the finer time instances was $\frac{1}{3}$, meaning 2 additional trajectory points were estimated between adjacent measurement times. Figure~\ref{random} displays one example of resulting networks.

The average TPR and PREC of the most likely networks over $100$ trials are recorded in Tables~\ref{T1}-\ref{T3}. GENIE3, dynGENIE3, ARNI  and BINGO are not included in these tables since they require a tuning threshold to select links. In all cases, our method considerably outperforms kernel-based approaches and GSBL. With high SNR ($20dB$), SDE\_TC and SDE\_DC present similar performance and are superior to the others.  Our method captures most true links even with a low number of data points ($TPR>90\%$). More importantly, PREC of SDE\_TC and SDE\_DC rises to $85\%$ as more data points are used. In contrast, PREC of kernel-based methods and GSBL are extremely low ($PREC<35\%$), meaning the inferred networks are not reliable.

As $SNR$ decreases to $10dB$, all methods require more data points to achieve accurate inference. The advantage of our method is evident compared with kernel-based methods and GSBL: TPR and PREC are always above $95\%$ and $70\%$, respectively. Although Kernel\_SS and Kernel\_DC present high TPR, their PREC is consistently below $55\%$, meaning around half of the inferred links are wrong.

It is remarkable that SDE\_TC and SDE\_DC are still able to provide reliable inference with high noise levels ($0dB$). Given $140$ data points, most true links in the ground truths are captured ($TPR>90\%$) and the inferred links are highly reliable ($PREC>80\%$). Similar to the $10dB$ case, PREC of kernel-based methods and GSBL are always below $50\%$.

Next, we compare our method with GENIE3, dynGENIE3, ARNI and BINGO by   calculating AUROC and AUPREC. For kernel-based methods and GSBL, the confidence of links is evaluated as $P(j\rightarrow i)=\frac{\|w_{ij}\|}{\|w_i\|}$~\cite{Aderhold2014}. Overall, our method considerably outperforms all the other methods. Due to the lack of space, only the result of $10dB$ is displayed in Table~\ref{T4}. It is clearly evident that the inferred networks of our method are highly reliable with both AUROC and AUPREC above $95\%$. GSBL, GENIE3, dynGENIE3 and ARNI show poor performance. In particular, their AUPREC are consistently below $35\%$. In addition, many true links are missed ($AUROC<70\%$). Kernel\_SS and Kernel\_DC cannot present good results unless at least $120$ time points are available.

The above results show that both PREC and AUPREC of other methods are considerably lower than ours. The poor performance of these methods is mainly caused by the low sampling frequency of measurements and partially observed networks. As GENIE3, GSBL and kernel-based methods require model discretization, a low sampling frequency changes ground truth topologies and produces fully-connected networks (basically, by inferring the topology of the exponential matrix of $A$, instead of the topology of $A$~\cite{yue}).  Additionally, a low sampling frequency leads to inaccurate estimate of derivatives, which degrades the performance of  ARNI and dynGENIE3. For BINGO, since only a partial set of nodes are measured it cannot infer the network topology ideally. However, our method can avoid these issues by refining sparsely measured time
series and using DSFs to describe the network topology of measured nodes. Hence, it is able to detect network topologies and identify internal dynamics accurately.

\begin{table}[h!]
	\caption{Inference of random networks with $20dB$ SNR}
	\centering
	\resizebox{\columnwidth}{!}{
		\begin{tabular}{ |c|c|c|c|c|c|c|}
			\hline
			\multicolumn{7}{|c|}{20dB}\\
			\hline
			\multirow{2}{2em}   &\multicolumn{2}{|c|}{40} &\multicolumn{2}{|c|}{60}&\multicolumn{2}{|c|}{80} \\
			\cline{2-7}
			&PREC &TPR&PREC&TPR&PREC&TPR\\
			\hline
		       GSBL      &14.9&37.9&13.4&52.8&14.2&68.0 \\
			Kernel\_TC      &  8.4	&4.5&15.0&11.5&28.5&54.0\\
			Kernel\_SS      &  15.6&59.5&17.2&83.7&	22.3	&93.8\\
			Kernel\_DC  	& 18.2&55.2&22.6&82.9&31.1&93.8\\
			SDE\_TC      &  66.5&92.7&79.5&95.2&85.4&98.2\\
			SDE\_SS      &  63.3&99.1&64.7&100&69.7&100\\
			SDE\_DC  &  63.7	&95.0&76.1&98.1&85.4	&98.6\\
			\hline
		\end{tabular}
	}
\label{T1}
\end{table}

\begin{table}[h!]
	\caption{Inference of random networks with $10dB$ SNR}
	\centering
	\resizebox{\columnwidth}{!}{
		\begin{tabular}{ |c|c|c|c|c|c|c|}
			\hline
			\multicolumn{7}{|c|}{10dB}\\
			\hline
			\multirow{2}{2em}   &\multicolumn{2}{|c|}{80} &\multicolumn{2}{|c|}{100}&\multicolumn{2}{|c|}{120}\\
			\cline{2-7}
			&PREC &TPR&PREC&TPR&PREC&TPR\\
			\hline
			 GSBL      &14.4&67.9&14.7&78.2&14.3&82.4 \\
			Kernel\_TC &21.0&15.3&31.6&23.4&54.5&23.4\\
			Kernel\_SS     &23.6	&90.2&35.8&93.7&42.5&98.0\\	
			Kernel\_DC     &29.6	&87.9&39.7&95.9&51.8&98.1\\		
			SDE\_TC      &	81.7&97.3&86.8&99.1&92.1&98.6\\
			SDE\_SS      &72.8	&100&71.2&100&72.8&100\\
			SDE\_DC     &81.6&96.9&83.9&99.1&89.5&98.5\\		
			\hline
		\end{tabular}
	}
\label{T2}
\end{table}
\begin{table}[h!]
	\caption{Inference of random networks with $0dB$ SNR}
	\centering
	\resizebox{\columnwidth}{!}{
		\begin{tabular}{ |c|c|c|c|c|c|c|}
			\hline
			\multicolumn{7}{|c|}{0dB}\\
			\hline
			\multirow{2}{2em}   &\multicolumn{2}{|c|}{100} &\multicolumn{2}{|c|}{120}&\multicolumn{2}{|c|}{140}\\
			\cline{2-7}
			&PREC &TPR&PREC&TPR&PREC&TPR\\
			\hline
			GSBL      &15.0&72.1&14.8&78.6&15.4&87.5\\
			Kernel\_TC      &25.7&30.4&31.0&33.7&36.5&46.8\\
			Kernel\_SS    	&32.7&85.2&40.2&94.7&48.2&96.2\\
			Kernel\_DC      &30.7	&86.0&35.3&88.1&48.2&92.5\\
			SDE\_TC      &76.6	&94.5&80.4&92.9&87.0&96.9\\
			SDE\_SS    	&60.1	&97.4&61.6&99.6&62.9&99.6\\
			SDE\_DC      &72.5	&90.6&77.4&93.2&81.9&93.5\\
			\hline
		\end{tabular}
	}
\label{T3}
\end{table}
\begin{table}[h!]
	\caption{Inference of random networks with $10dB$ SNR}
	\centering
	\resizebox{\columnwidth}{!}{
		\begin{tabular}{ |c|c|c|c|c|c|c|}
			\hline
			\multicolumn{7}{|c|}{10dB}\\
			\hline
			\multirow{2}{2em}   &\multicolumn{2}{|c|}{80} &\multicolumn{2}{|c|}{100}&\multicolumn{2}{|c|}{120}\\
			\cline{2-7}
			&AUPREC &AUROC&AUPREC&AUROC&AUPREC&AUROC\\
			\hline
			ARNI &  16.9	&66.2&18.9&69.7&18.2	&66.8\\
			dynGENIE3 &  13.5	&44.0&13.8&47.4&14.5	&45.9\\
			GENIE3 & 33.4 	&68.0&32.0&67.3&34.8&68.9\\
			 GSBL      &18.7&58.3&22.2&64.1&25.8&64.8 \\
			BINGO      &56.6&82.4&63.1&86.8&69.1&88.3 \\
			Kernel\_TC &22.6&54.2&33.0&57.9&47.4&60.0\\
			Kernel\_SS &65.7&86.9&85.1&94.3&92.3&97.7\\
			Kernel\_DC &68.9&88.3&81.3&94.4&91.1&97.5\\
			SDE\_TC      &96.9	&98.7&97.6&99.2&98.0&99.0\\
			SDE\_SS      &96.9	&98.7&99.6&99.9&99.6&99.9\\
			SDE\_DC     &96.6&98.9&97.8&99.2&97.7&99.0\\		
			\hline
		\end{tabular}
	}
	\label{T4}
\end{table}

\subsection{Ring networks}

\begin{figure}
	\centering
	\includegraphics[width=0.5\columnwidth]{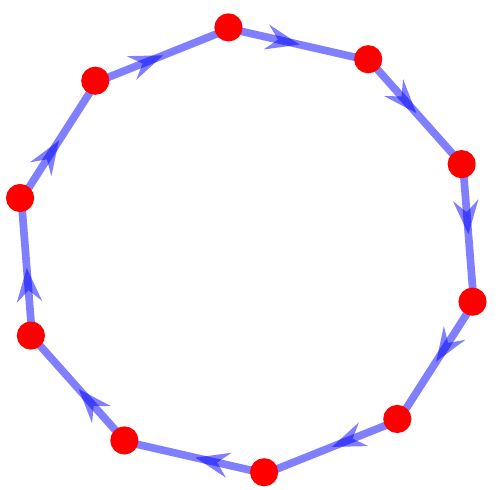}
	\caption{A ring network of measured nodes.}
	\label{ring} 
\end{figure}

$100$ networks with the fixed ring structure as shown in Figure~\ref{ring} were generated and simulated following the same protocol in the last section. Since ring networks contain a feedback loop and are extremely sparse, they are more challenging to infer.

Table~\ref{T5} presents the inference result. Our method is able to capture most true links ($TPR>95\%$). More importantly, PREC of SDE\_TC and SDE\_DC approaches $90\%$ as more data points are used for inference. In contrast, PREC of kernel-based methods and GSBL are extremely low, resulting in unreliable networks.

Table~\ref{T6} indicates that our method also outperforms BINGO, GENIE3, dynGENIE3 and ARNI. In particular, AUPREC of these methods are always below $60\%$, meaning that inferred networks are not sparse, compared with the ring structure, and most inferred links are wrong.

\begin{table}[h!]
	\caption{Inference of ring networks with $10dB$ SNR}
	\centering
	\resizebox{\columnwidth}{!}{
		\begin{tabular}{ |c|c|c|c|c|c|c|}
			\hline
			\multicolumn{7}{|c|}{10dB}\\
			\hline
			\multirow{2}{2em}   &\multicolumn{2}{|c|}{60} &\multicolumn{2}{|c|}{80}&\multicolumn{2}{|c|}{100}\\
			\cline{2-7}
			&PREC &TPR&PREC&TPR&PREC&TPR\\
			\hline
						GSBL             &13.0&51.8&14.1&80.5&13.7 &78.5\\
			Kernel\_TC   &15.0&10.4&16.0&14.0&37.4&17.5\\
			Kernel\_SS     &18.2&73.7&24.5&91.9&30.5&98.2\\
			Kernel\_DC     &25.2&78.1&30.4&91.7&44.7&95.2\\	
			SDE\_TC      &70.1&98.5&83.1&99.0&88.8&100\\
			SDE\_SS      &61.9&99.5&69.9&99.5&73.7&100\\
			SDE\_DC     &68.0&98.0&84.5&99.5&85.0&98.5\\	
			\hline
		\end{tabular}
	}
\label{T5}
\end{table}

\begin{table}[h!]
	\caption{Inference of ring networks with $10dB$ SNR}
	\centering
	\resizebox{\columnwidth}{!}{
		\begin{tabular}{ |c|c|c|c|c|c|c|}
			\hline
			\multicolumn{7}{|c|}{10dB}\\
			\hline
			\multirow{2}{2em}   &\multicolumn{2}{|c|}{60} &\multicolumn{2}{|c|}{80}&\multicolumn{2}{|c|}{100}\\
			\cline{2-7}
			&AUPREC &AUROC&AUPREC&AUROC&AUPREC&AUROC\\
			\hline
			ARNI      &16.5&62.4&18.1&66.4&17.7&67.3\\
			dynGENIE3 &15.1&46.9&16.0&43.0&15.7&45.4\\
			GENIE3   &29.7&66.7&30.3&66.4&34.6&67.6\\
			GSBL    &16.0 &54.4&18.7&60.2 &22.6&62.9 \\
			BINGO    &47.2 &79.7&57.5&84.2 &58.7&83.3 \\
			Kernel\_TC &15.0&51.5&20.2&52.9&34.1&56.4\\
			Kernel\_SS &39.6&71.7&67.3&88.3&82.3&95.4\\
			Kernel\_DC &50.2&78.8&64.5&88.2&77.5&93.8\\
			SDE\_TC      &97.7&99.2&97.5	&99.5&98.6&99.8\\
			SDE\_SS      &99.2&99.9&99.6&99.9&99.4&99.9\\
			SDE\_DC     &94.9&98.5&98.5	&99.8&98.0&99.7\\		
			\hline
		\end{tabular}
	}
\label{T6}
\end{table}

\subsection{Synthetic networks}
We tested all methods on a synthetic biological network of yeast \textit{Saccharomyces cerevisiae}~\cite{yeast}. Data from this artificial network was obtained from real-life biochemical experiments. Hence, the working mechanisms of such a network are expected to be considerably more complex than linear networks. Whilst topologies of most practical biological networks remain unknown, the topology of this synthetic network is known \textit{a priori}, which makes it suitable to test different methods. The network contains $5$ nodes and measurements were available for all nodes. Time series were collected with different lengths (i.e. $15$, $19$ and $21$). Since the sampling frequency was consistent, we set up a pseudo-time with sampling frequency $1$.  To refine measurements, the increment step ($\varDelta T$) of finer time instances was $\frac{1}{3}$. 

The inference result is displayed in Table~\ref{T7}. Our method considerably outperforms all other methods. AUROC and AUPREC of our method are always above $64\%$ whilst those of the other methods are very low in most cases. The result indicates that our method has great potential to infer real-world networks.
\begin{table}[h!]
	\caption{Inference of synthetic networks}
	\centering
	\resizebox{\columnwidth}{!}{
		\begin{tabular}{ |c|c|c|c|c|c|c|}
			\hline
			\multicolumn{7}{|c|}{10dB}\\
			\hline
			\multirow{2}{2em}   &\multicolumn{2}{|c|}{15} &\multicolumn{2}{|c|}{19}&\multicolumn{2}{|c|}{21}\\
			\cline{2-7}
			&AUPREC &AUROC&AUPREC&AUROC&AUPREC&AUROC\\
			\hline
			ARNI &47.7&46.9&49.3&56.7&42.0&54.2\\
			dynGENIE3 &43.3&39.6&56.0&50.0&49.2&63.5\\
			GENIE3 &47.7&61.5&49.3&51.0&42.0&57.3\\
			GSBL  &52.3&60.8&36.3&44.5&52.0&61.0\\
			BINGO  &71.6&72.9&54.4&62.7&55.8&60.4\\
			Kernel\_TC &34.2&46.9&65.1&61.5&70.0&50.0\\
			Kernel\_SS &29.8&37.5&44.8&58.3&40.1&51.4\\
			Kernel\_DC &41.2&44.3&75.7&79.2&53.5&55.2\\
			SDE\_TC      &68.3&64.6&87.7&85.4&78.7&71.9\\	
			SDE\_SS      &71.2&69.8&85.0&83.3&65.8&69.8\\
			SDE\_DC     &78.1&75.5&70.1&75.5&73.9&71.9\\		
			\hline
		\end{tabular}
	}
\label{T7}
\end{table}

\section{CONCLUSION AND DISCUSSION}\label{Conclusion}

This paper presented a data-driven method to infer linear stochastic dynamical networks. Continuous-time networks are described by linear SDE models with unmeasured nodes encoded via transfer functions. MCMC techniques are used to explore network topologies and internal dynamics. In particular, the pCN sampling scheme is adopted to refine sparsely sampled time series. The estimated trajectories greatly improve inference accuracy. Monte Carlo simulations imply that our method considerably outperforms several state-of-the-art methods including kernel-based methods, GSBL, BINGO, GENIE3, dynGENIE3 and ARNI under different experimental conditions. Whilst the proposed method is developed for linear networks, simulations on a synthetic biological network indicate that our method is also capable to deal with real-world networks.

Overall, the value of this approach is that it can tackle poor measurements collected with a low sampling frequency, varying sampling steps and measurement noise.  By exploring trajectories at finer time instances, our method is able to refine poor measurements, thus ensuring robust performance. The main drawback of our method is heavy computational cost. Although a matrix approximation scheme is embedded in our framework to save computational load, the growth of cost is still cubic with respect to the size of networks.
Future developments should focus on reducing computational cost of the algorithm so that the proposed method can be applied to large scale networks. More importantly, it is necessary to extend our method to nonlinear networks.
\ifCLASSOPTIONcaptionsoff
  \newpage
\fi



%
 \bibliographystyle{unsrt}        
	\bibliography{autosam} 




%








\end{document}